\documentclass[10pt,twocolumn,letterpaper]{article}

\usepackage[pagenumbers]{cvpr} %

\usepackage{epsfig}
\usepackage{graphicx}
\usepackage{amsmath}
\usepackage{amssymb}
\usepackage{booktabs}

\usepackage[pagebackref,breaklinks,colorlinks]{hyperref}

\usepackage[capitalize]{cleveref}
\crefname{section}{Sec.}{Secs.}
\Crefname{section}{Section}{Sections}
\Crefname{table}{Table}{Tables}
\crefname{table}{Tab.}{Tabs.}

\def\cvprPaperID{1800} %
\def\confName{CVPR}
\def\confYear{2022}
\makeatletter
\apptocmd\@maketitle{{\teaserfigure{}\par}}{}{}
\makeatother
\newcommand{\teasernobox}{
\begingroup
\begin{subfigure}[b]{0.170\linewidth}
	    \centering
    	\includegraphics[width=\linewidth]{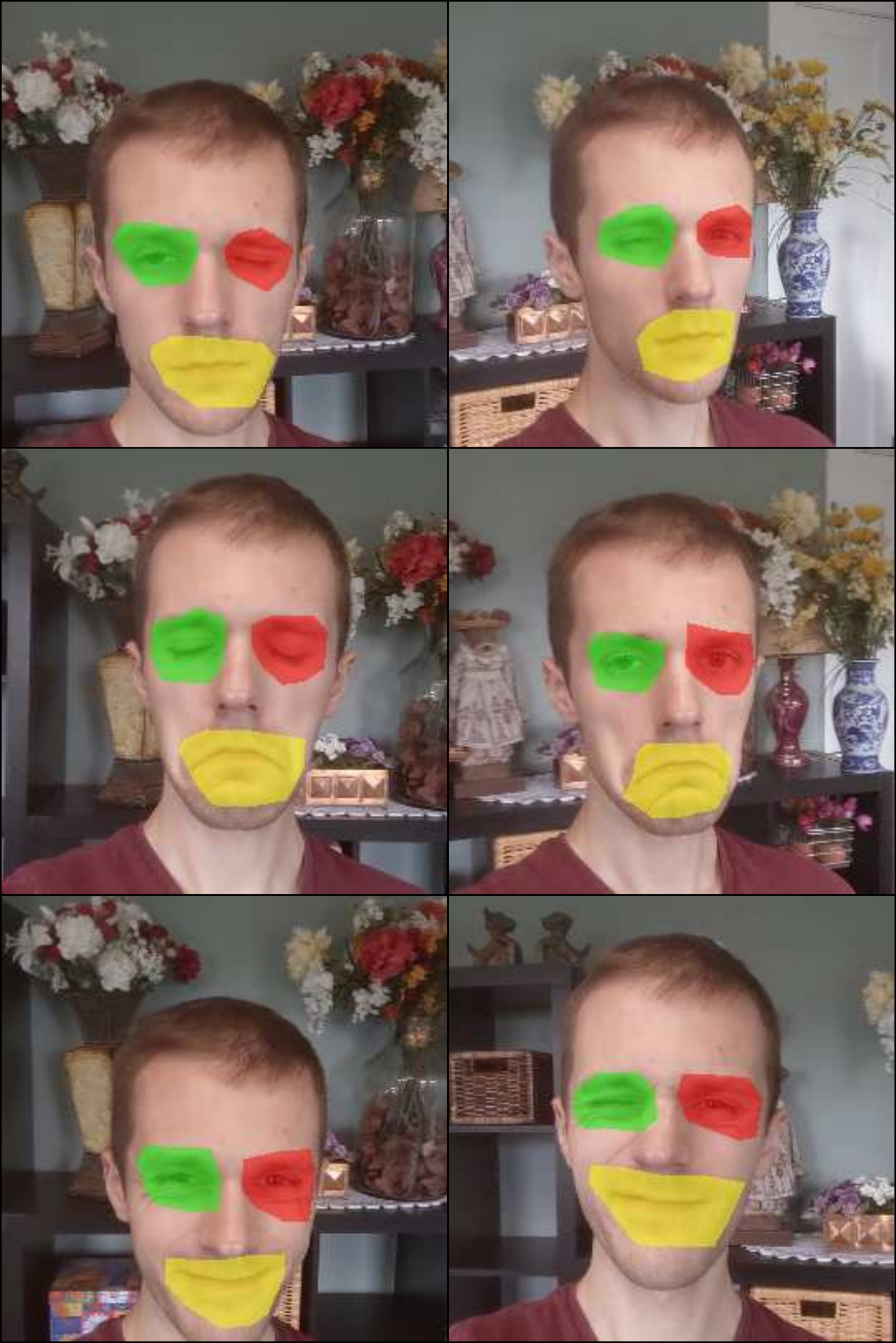}
		\subcaption{annotations}
\end{subfigure}
\hfill{}
\begin{subfigure}[b]{0.42\linewidth}
	    \centering
    	\includegraphics[width=\linewidth]{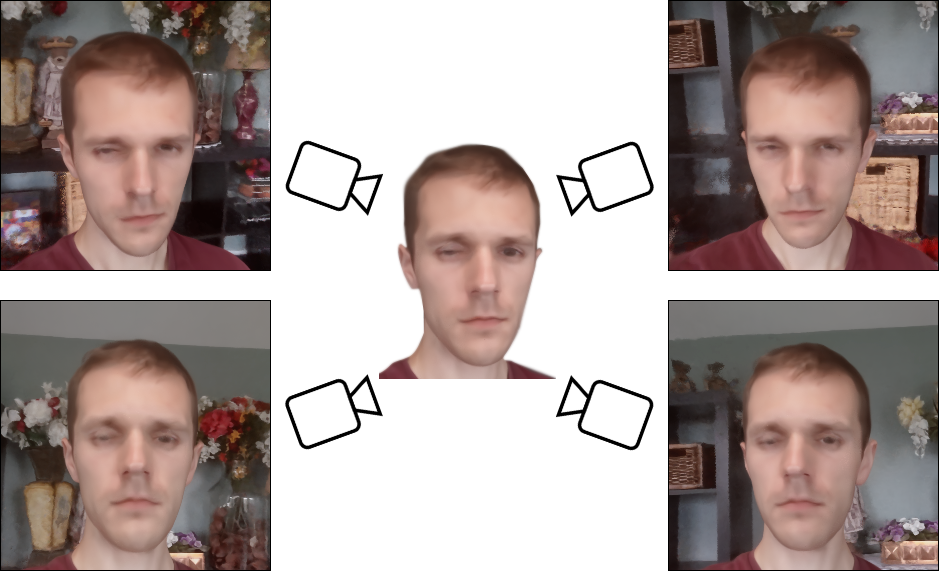}
		\subcaption{novel views}
\end{subfigure}
\hfill{}
\begin{subfigure}[b]{0.374\linewidth}
	    \centering
    	\includegraphics[width=\linewidth]{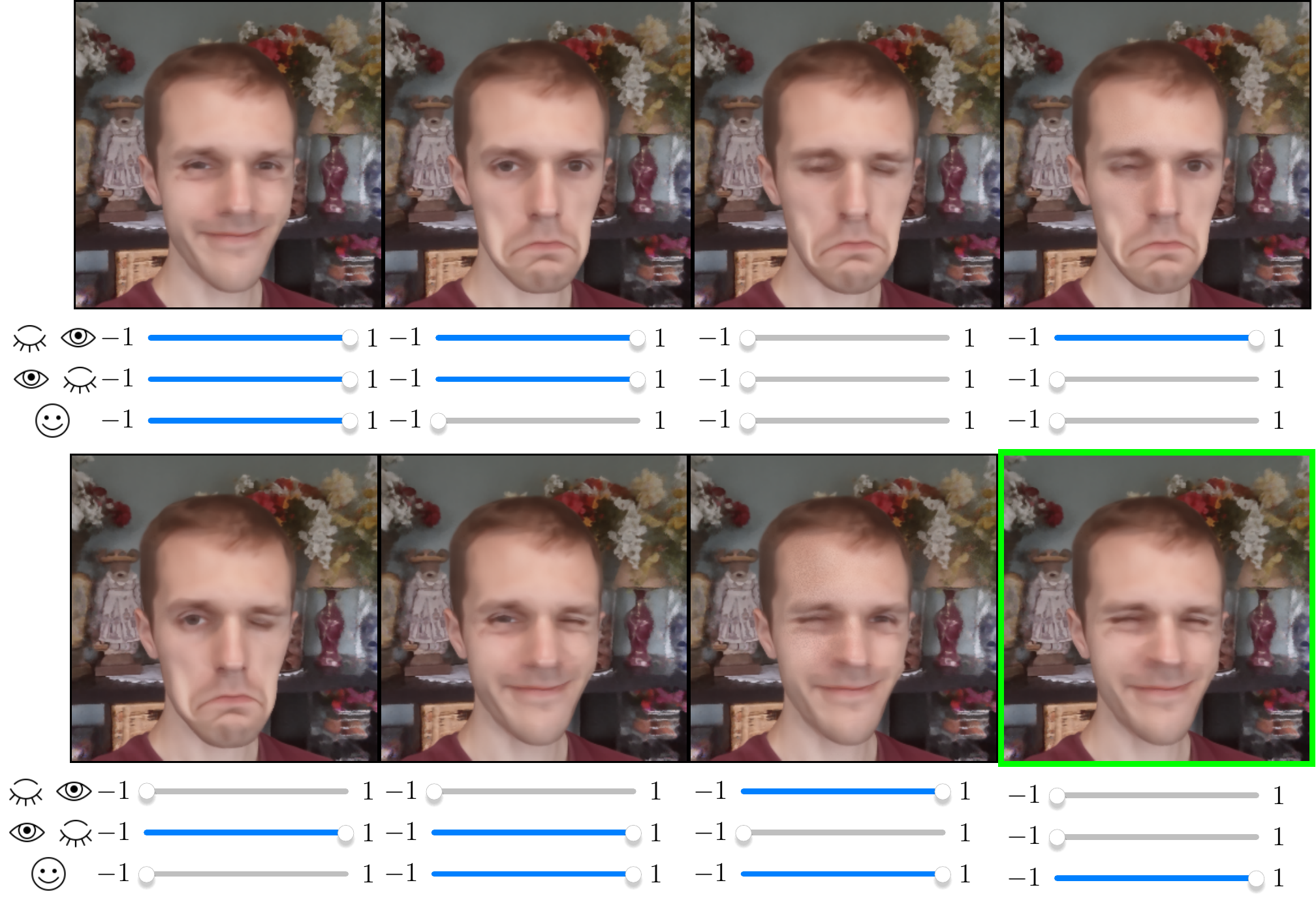}
		\subcaption{novel attributes}
\end{subfigure}
\endgroup
}

\newcommand{\teaserfigure}{
\vspace{-5mm}
\captionsetup[sub]{labelformat=parens}
\captionsetup{type=figure}

\teasernobox

\vspace{-2mm}
\setcounter{figure}{0} %
\captionsetup{type=figure}
\captionof{figure}{
{\bf Teaser --}
We train a controllable neural radiance field from multiple views of a dynamic 3D scene, under varying poses and attributes; in this example eye being open/closed and mouth smiling/frowning.
Given only six annotations (a), our method provides full control over the scene appearance, allowing us to synthesize (b) novel views and (c) novel attributes, including attribute combinations that were \textit{never seen} in the training data~(green box).
}
\vspace{4mm}
\label{fig:teaser}
}

\usepackage[inline]{enumitem} %
\usepackage{color}
\usepackage{multirow}
\usepackage[makeroom]{cancel}
\usepackage{placeins} %

\definecolor{turquoise}{cmyk}{0.65,0,0.1,0.3}
\definecolor{purple}{rgb}{0.65,0,0.65}
\definecolor{dark_green}{rgb}{0, 0.5, 0}
\definecolor{orange}{rgb}{0.8, 0.6, 0.2}
\definecolor{red}{rgb}{0.8, 0.2, 0.2}
\definecolor{darkred}{rgb}{0.6, 0.1, 0.05}
\definecolor{blueish}{rgb}{0.0, 0.3, .6}
\definecolor{light_gray}{rgb}{0.7, 0.7, .7}
\definecolor{pink}{rgb}{1, 0, 1}
\definecolor{greyblue}{rgb}{0.25, 0.25, 1}

\newcommand{\comment}[1]{}

\newcommand{\ethicspart}[1]{#1}

\newcommand{\CIRCLE}[1]{\raisebox{.5pt}{\footnotesize \textcircled{\raisebox{-.6pt}{#1}}}}

\newcommand{\Figure}[1]{Figure~\ref{fig:#1}}

\newcommand{\Table}[1]{Table~\ref{tab:#1}}
\newcommand{\eq}[1]{\eqref{eq:#1}}
\newcommand{\Eq}[1]{Eq.~\eqref{eq:#1}}

\newcommand{\Section}[1]{Section~\ref{sec:#1}}

\usepackage{blindtext}

\def \customparskip {.5em}
\renewcommand{\paragraph}[1]{\vspace{\customparskip}\noindent\textbf{#1}.}

\newenvironment{rcases}
  {\left.\begin{aligned}}
  {\end{aligned}\right\rbrace}

\usepackage{dsfont}
\DeclareMathOperator*{\argmin}{arg\,min}

\newcommand{\loss}[1]{\mathcal{L}_\text{#1}}

\newcommand{\expect}{\mathbb{E}}
\newcommand{\real}{\mathbb{R}}

\newcommand{\balpha}{{\boldsymbol{\alpha}}}
\newcommand{\bbeta}{{\boldsymbol{\beta}}}

\newcommand{\bx}{{\mathbf{x}}}
\newcommand{\bc}{{\mathbf{c}}}

\newcommand{\bC}{{\mathbf{C}}}

\newcommand{\br}{{\mathbf{r}}}
\newcommand{\bv}{{\mathbf{v}}}

\newcommand{\Canonicalizer}{{\mathcal{K}}}
\newcommand{\hypermap}{\mathcal{H}}
\newcommand{\Representation}{{\mathcal{R}}}
\newcommand{\Attribute}{{\mathcal{A}}}

\newcommand{\pars}{\boldsymbol{\theta}}
\newcommand{\given}{~|~}

\newcommand{\iImage}{c}
\newcommand{\nImages}{C}
\newcommand{\image}{\mathbf{C}}
\newcommand{\images}{\{\image_\iImage\}}
\newcommand{\latents}{\{\boldsymbol\beta_\iImage\}}
\newcommand{\latent}{\boldsymbol\beta}
\newcommand{\latentDim}{B}

\newcommand{\allpars}{\boldsymbol\theta}

\newcommand{\width}{W}
\newcommand{\height}{H}
\newcommand{\nAttributes}{A}
\newcommand{\hyperAttributeDim}{d}
\newcommand{\iAttribute}{a}
\newcommand{\attribute}{\alpha}
\newcommand{\attributes}{\boldsymbol\alpha}
\newcommand{\Mask}{\mathbf{M}}
\newcommand{\mask}{\mathbf{m}}
\newcommand{\indicator}{\delta}
\newcommand{\gt}{\text{gt}}
\newcommand{\AttributeNet}{\mathcal{A}}
\newcommand{\point}{\mathbf{x}}
\newcommand{\MaskNet}{\mathcal{M}}

\newcommand{\ray}{\br}

\newcommand{\SupplementaryMaterial}{\texttt{Supplementary Material}\xspace}
\begin{document}
\title{CoNeRF: Controllable Neural Radiance Fields}

\author{
Kacper Kania\textsuperscript{1,2} \quad
Kwang Moo Yi\textsuperscript{1} \quad
Marek Kowalski\textsuperscript{4} \quad
Tomasz Trzciński\textsuperscript{2} \quad
Andrea Tagliasacchi\textsuperscript{3,5} \quad
\\[.5em]
University of British Columbia\textsuperscript{1} \quad
Warsaw University of Technology\textsuperscript{2} \quad 
University of Toronto\textsuperscript{3} 
\\
Microsoft\textsuperscript{4} \quad
Google Research\textsuperscript{5}
}
\maketitle
\begin{abstract}
We extend neural 3D representations to allow for intuitive and interpretable user control beyond novel view rendering (i.e. camera control). We allow the user to annotate which part of the scene one wishes to control with just a small number of mask annotations in the training images. Our key idea is to treat the attributes as latent variables that are regressed by the neural network given the scene encoding. This leads to a few-shot learning framework, where attributes are discovered automatically by the framework, when annotations are not provided. We apply our method to various scenes with different types of controllable attributes (e.g. expression control on human faces, or state control in movement of inanimate objects). Overall, we demonstrate, to the best of our knowledge, for the first time novel view and novel attribute re-rendering of scenes from a single video.
\end{abstract}

\section{Introduction}
\label{sec:intro}
Neural radiance field (NeRF)~\cite{mildenhall2020nerf} methods have recently gained popularity thanks to their ability to render photorealistic novel-view images~\cite{martin2021nerf, park2020deformable,park2021hypernerf,zhang2020nerf++}.
In order to widen the scope to other possible applications, such as digital media production, a natural question is whether these methods could be extended to enable \textit{direct} and \textit{intuitive} control by a digital artist, or even a casual user.
However, current techniques only allow coarse-grain controls over materials~\cite{zhang2021nerfactor}, color~\cite{jang2021codenerf}, or object placement~\cite{yang2021learning}, or only support changes that they are designed to deal with, such as shape deformations on a learned shape space of chairs~\cite{liu2021editing}, or are limited to facial expressions encoded by an explicit face model~\cite{gafni2021dynamic}.
By contrast, we are interested in \textit{fine-grained} control without limiting ourselves to a specific class of objects or their properties.
For example, given a self-portrait video, we would like to be able to control individual \textit{attributes} (e.g. whether the mouth is open or closed); see~\Figure{teaser}.
We would like to achieve this objective with minimal user intervention, without the need of specialized capture setups~\cite{liu2021neural}.

However, it is unclear how fine-grained control can be achieved,
as current state-of-the-art models~\cite{park2021hypernerf} encode the structure of the 3D scene in a \textit{single} and \textit{not interpretable} latent code.
For the example of face manipulation, one could attempt to resolve this problem by providing \textit{dense} supervision by matching images to the corresponding Facial Action Coding System (FACS)~\cite{facs} action units.
Unfortunately, 
this would require either an automatic annotation process or careful and extensive per-frame human annotations, making the process expensive, generally unwieldy, and, most importantly, domain-specific.
Automated tools for domain-agnostic latent disentanglement are a very active topic of research in machine learning~\cite{higgins2016beta, higgins2018towards, chen2016infogan}, but no effective plug-and-play solution exists yet.

Conversely, we borrow ideas from 3D morphable models~(3DMM)~\cite{blanz1999morphable}, and in particular to recent extensions that achieve local control by \textit{spatial disentanglement} of control attributes~\cite{wu2016anatomically,neumann2013sparse}.
Rather than having a single global code controlling the expression of the \textit{entire} face, we would like to have a set of \textit{local} ``attributes'', each controlling the corresponding \textit{localized} appearance; more specifically, we assume spatial quasi-conditional independence of attributes~\cite{wu2016anatomically}.
For our example in~\Figure{teaser}, we seek an attribute capable to control the appearance of the mouth, another to control the appearance of the eye, etc.

Thus, we introduce a learning framework denoted CoNeRF~(i.e. Controllable~NeRF) that is capable of achieving this objective with just \textit{few-shot} supervision.
As illustrated in \Figure{teaser}, given a single one-minute video, and with as little as two annotations per attribute, CoNeRF allows \textit{fine-grained}, \textit{direct}, and \textit{interpretable} control over attributes.
Our core idea is to provide, on top of the ground truth attribute tuple, \textit{sparse} 2D mask annotations that specify which region of the image an attribute controls.
Further, by treating attributes as latent variables within the framework, the mask annotations can be automatically propagated to the whole input video.
Thanks to the quasi-conditional independence of attributes, our technique allows us to synthesize expressions that were \textit{never} seen at training time; e.g.~the input video never contained a frame where both eye were closed and the actor had a smiling expression; see~\Figure{teaser}~(green box).

\paragraph{Contributions}
To summarize, our CoNeRF method\footnote{Code and dataset will be released if the paper is accepted.}:
\vspace{-.5em}
\begin{itemize}[leftmargin=*]
\setlength\itemsep{-.3em}
\item provides \textit{direct}, \textit{intuitive}, and \textit{fine-grained} control over 3D neural representations encoded as NeRF;
\item achieves this via \textit{few-shot} supervision, \eg, just a handful of annotations in the form of attribute values and corresponding 2D mask are needed for a one minute video;
\item while inspired by domain-specific facial animation research~\cite{wu2016anatomically}, it provides a \textit{domain-agnostic} technique.
\end{itemize}

\color{black}

\section{Related works}
\label{sec:related}
Neural Radiance Fields~\cite{mildenhall2020nerf} provide high-quality  renderings of scenes from novel views with just a few exemplar images captured by a handheld device.
Various extensions have been suggested to date.
These include ones that focus on improving the quality of results~\cite{martin2021nerf, park2020deformable, park2021hypernerf, zhang2020nerf++}, ones that allow a single model to be used for multiple scenes~\cite{schwarz2020graf, trevithick2020grf}, and some considering
controllability of the rendering output at a coarse level~\cite{guo2020object, yu2021unsupervised, liu2021editing, yang2021learning, xie2021fig, zhang2021nerfactor}, as we detail next.

In more detail, existing works enable only compositional control of object location~\cite{yang2021learning,yu2021unsupervised}, and recent extensions also allow for finer-grain reproduction of global illumination effects~\cite{guo2020object}.
NeRFactor~\cite{zhang2021nerfactor} shows one can model albedos and BRDFs, and shadows, which can be used to, \eg, edit material, but the manipulation they support is limited to what is modeled through the rendering equation.
CodeNeRF~\cite{jang2021codenerf} and EditNeRF~\cite{liu2021editing} showed that one can edit NeRF models by modifying the shape and appearance encoding, but they require a curated dataset of objects viewed under different views and colors.
HyperNeRF~\cite{park2021hypernerf}, on the other hand can adapt to unseen changes specific to the scene, but learns an arbitrary attribute (ambient) space that cannot be supervised, and, as we show in \Section{results}, cannot be easily related to specific local attribute within the scene for controllability.

\paragraph{Explicit supervision}
One can also condition NeRF representations~\cite{gafni2021dynamic} with face attribute predicted by pre-trained face tracking networks, such as Face2Face~\cite{thies2016face2face}.
Similarly, for human bodies, A-NeRF~\cite{su2021anerf} and NARF~\cite{noguchi2021neural} use the SMPL~\cite{loper2015smpl} model to generate interpretable pose parameters, and Neural~Actor~\cite{liu2021neural} further includes normal and texture maps more detailed rendering.
While these models result in controllable NeRF, they are limited to domain-specific control and the availability of a heavily engineered control model.

\paragraph{Controllable neural implicits}
Controllability of neural 3D \textit{implicit} representations has also been addressed by the research community.
Many works have limited focus on learning \textit{human} neural implicit representations while enabling the control via SMPL parameters~\cite{loper2015smpl}, or linear blend skinning weights~\cite{zheng2021pamir, he2021arch++, saito2021scanimate, mihajlovic2021leap, ma2021scale, deng2020nasa, alldieck2021imghum, zins2021data}. Some initial attempts at learned disentangled of shape and poses have also been made in A-SDF~\cite{mu2021sdf}, allowing behavior control of the output geometry~(\eg doors open vs. closed) while maintaining the general shape. However, the approach is limited to controlling $\text{SE}(3)$ articulation of objects, and requires dense 3D supervision.

\subsection{Neural Radiance Field (NeRF)}
For completeness, we briefly discuss NeRF before diving into the details of our method.
A Neural Radiance Field captures a volumetric representation of a specific scene within the weights of a neural network.
As input, it receives a sample position~$\bx$ and a view direction~$\bv$ and outputs the density of the scene $\sigma$ at position~$\bx$ as well as the color~$\bc$ at position~$\bx$ as seen from view direction~$\bv$.
One then renders image pixels~$\bC$ via volume rendering~\cite{kajiya1984ray}.
In more detail, $\bx$ is defined by observing rays~$\br(t)$ as \mbox{$\bx=\br(t)$}, where $t$ parameterizes at which point of the ray you are computing for.
One then renders the color of each pixel $\bC(\br)$ by computing 
\begin{equation}
    \bC\left(\br\right) = \int_{t_n}^{t_f} T(t) \sigma\left(\br(t)\right) \bc\left(\br(t), \bv\right) dt
    \;,
    \label{eq:volume_render}
\end{equation}
where $\bv$ is the viewing angle of the ray $\br$, $t_n$ and $t_f$ are the near and far planes of the rendering volume, and 
\begin{equation}
    T(t) = \exp \left ( - \int_{t_n}^t \sigma({\bf r}(s)) ds \right )
    \;,
\end{equation}
is the accumulated transmittance.
Integration in \eq{volume_render} is typically done via numerical integration~\cite{mildenhall2020nerf}.

\subsection{HyperNeRF}
Note that in its original formulation~\eq{volume_render} is only able to model \textit{static} scenes.
Various recent works~\cite{park2020deformable, park2021hypernerf, tretschk2021non} have been proposed to explicitly account for possible appearance changes in a scene (for example, temporal changes in a video).
To achieve this, they introduce the notion of \textit{canonical} \textit{hyperspace} -- more formally given a 3D query point $\bx$ and the collection $\pars$ of all parameters that describe the model, they define:
\begin{align}
\Canonicalizer(\bx) &\equiv \Canonicalizer(\bx \given \bbeta, \pars ), \quad&\text{Canonicalizer} \label{eq:canonicalizer}\\
\bbeta(\bx) &\equiv \hypermap (\bx \given \bbeta, \pars ), \quad&\text{Hyper~Map} \label{eq:hypermap}\\
\bc(\point), \sigma(\point) &= \Representation(\Canonicalizer(\bx), \bbeta(\bx) \given \pars). \quad&\text{Hyper~NeRF}
\label{eq:hypernerf}
\end{align}
where the location is canonicalized via a canonicalizer $\Canonicalizer$, and the appearances, represented by $\bbeta$, are mapped to a hyperspace via $\hypermap$, which are then utilized by another neural network $\Representation$ to retrieve the color $\bc$ and the density $\sigma$ at the query location.
Note throughout this paper we denote $\latent$ to indicates a latent code, while $\latent(\point)$ to indicate the corresponding field generated by the hypermap lifting.
With this latent lifting, these methods render the scene via \Eq{volume_render}.
Note that the original NeRF model can be thought of the case where $\Canonicalizer$ and $\hypermap$ are identity mappings.

\begin{figure*}
\centering
\begin{subfigure}[b]{0.39\linewidth}
    \centering
    \includegraphics[width=\linewidth]{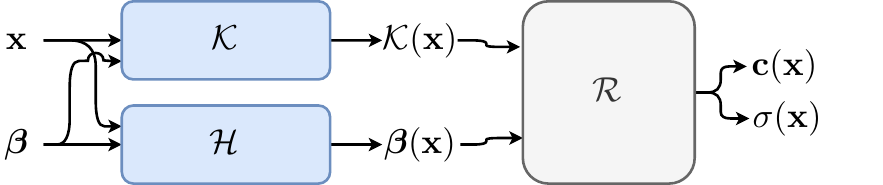}
    \vspace{0.0em}
    \caption{Neural radiance field with lifting~\cite{park2021hypernerf}}
    \label{fig:implicit}
\end{subfigure}
\hfill
\begin{subfigure}[b]{0.60\linewidth}
    \centering
    \includegraphics[width=\linewidth]{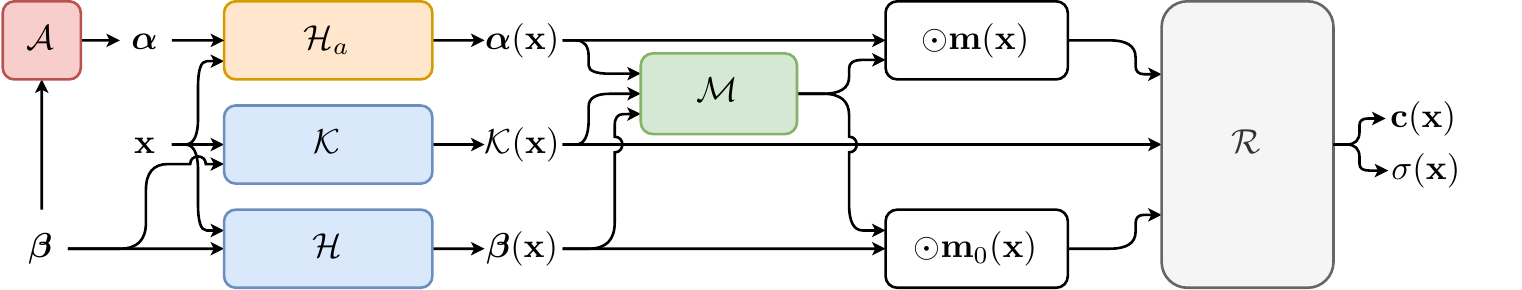}
    \caption{Controllable neural radiance field (our method)}
    \label{fig:laced-implicit}
\end{subfigure}
\caption{{\bf Framework -- }
We depict in (a) the HyperNeRF~\cite{park2021hypernerf} formulation, and (b) our Controllable-NeRF~(CoNeRF).
In (a), both point coordinates $\mathbf{x}$ and latent representation $\boldsymbol{\beta}$ are respectively processed by a canonicalizer $\Canonicalizer$ and a hyper map $\hypermap$, which are then turned into radiance and density field values by $\Representation$.
In (b), we introduce regressors $\Attribute$ and $\MaskNet$ that regress the attribute and the corresponding mask that enable few-shot attribute-based control of the NeRF model.
See~\Section{inference} for details.
}
\label{fig:pipeline}
\end{figure*} 
\section{Controllable NeRF (CoNeRF)}
Given a collection of $\nImages$ color images $\images \in [0,1]^{\width \times \height \times 3}$, we train our controllable neural radiance field model by an auto-decoding optimization~\cite{park2019deepsdf} whose losses can be grouped into two main subsets:
\begin{align}
\argmin_{\allpars=\pars, \latents} 
\underbrace{\loss{rep}(\allpars \given \images)}_\text{\Section{autodecode}}{+}
\underbrace{\loss{ctrl}(\allpars \given
\{\Mask^\gt_{\iImage, \iAttribute} \}, \{\attribute^\gt_{\iImage,\iAttribute}\} )}_\text{\Section{control}}
.
\end{align}
The first group consists of the classical HyperNeRF~\cite{park2021hypernerf} auto-decoder losses, attempting to optimize neural network parameters $\pars$ jointly with latent codes $\latents$ to \textit{reproduce} the corresponding input images $\images$:
\begin{align}
\loss{rep}(\cdot) =
\loss{recon}(\pars, \latents \given \images) +
\loss{enc}(\latents)
.
\label{eq:loss_render}
\end{align}
The latter allow us to inject \textit{explicit control} into the representation, and are our core contribution: 
\begin{align}
\loss{ctrl}(\cdot) 
&= \loss{mask}(\pars, \latents \given \{\Mask^\gt_{\iImage, \iAttribute} \}) &\text{g.t. masks}\\
&+ \loss{attr}(\pars, \latents \given \{\attribute^\gt_{\iImage,\iAttribute}\}). &\text{g.t. attributes}
\label{eq:loss_control}
\end{align}
As mentioned earlier in \Section{intro}, we aim for a 
neural 3D appearance model that is controlled by a collection of attributes $\attributes {=} \{\attribute_\iAttribute\}$, and we expect each image to be a manifestation of a different value of attributes, that is, each image $\image_\iImage$, and hence each latent code $\latent_\iImage$, will have a corresponding attribute~$\attributes_\iImage$.
The learnable connection between latent codes $\latent$ and the attributes $\attributes$, 
which we represent via regressors, is detailed in~\Section{inference}.

\subsection{Reconstruction losses}
\label{sec:autodecode}
The primary loss guiding the training of the NeRF model is the reconstruction loss, which simply aims to reconstruct observations $\images$. 
As in other neural radiance field models~\cite{mildenhall2020nerf, martin2021nerf, park2020deformable, park2021hypernerf} we simply minimize the L2 photometric reconstruction error with respect to ground truth images:
\begin{equation}
\loss{recon}(\cdot) = \sum_\iImage \expect_{\ray \sim \image_\iImage} \left[ \left\|\bC(\ray \given \latent_\iImage, \pars)-\bC^\text{gt}(\ray) \right\|_2^2 \right]
.
\label{eq:recon}
\end{equation}
As is typical in auto-decoders, and following~\cite{park2019deepsdf}, we impose a zero-mean Gaussian prior on the latent codes~$\latents$:
\begin{equation}
\loss{enc}(\cdot) = \sum_\iImage \left\| \latent_\iImage \right\|^2_2
.
\label{eq:latent_prior}
\end{equation}

\subsection{Control losses}
\label{sec:control}
The user defines a \textit{discrete} set of $\nAttributes$ number of attributes that they seek to control, that are \textit{sparsely} supervised across frames---we only supervise attributes \textit{when} we have an annotation, and let others be discovered on their own throughout the training process, as guided by~\eq{loss_render}.
More specifically, for a particular image $\image_\iImage$, and a particular attribute $\attribute_\iAttribute$, the user specifies the quantities:
\begin{itemize}
\item $\attribute_{\iImage, \iAttribute} \in [-1,1]$: specifying the value for the $\iAttribute$-th attribute in the $\iImage$-th image; see the \textit{sliders} in~\Figure{teaser};
\item $\Mask_{\iImage, \iAttribute} \in [0,1]^{\width \times \height}$: roughly specifying the image region that is controlled by the $\iAttribute$-th attribute in the $\iImage$-th image; see the \textit{masks} in~\Figure{teaser}.
\end{itemize}
To formalize sparse supervision, we employ an indicator function $\indicator_{\iImage, \iAttribute}$,
where $\indicator_{\iImage, \iAttribute}=1$ if 
 an annotation for attribute $\iAttribute$ for image $\iImage$ is provided, otherwise $\indicator_{\iImage, \iAttribute}=0$.
We then write the loss for \textit{attribute} supervision as:
\begin{equation}
\loss{attr}(\cdot) =
\sum_\iImage \sum_\iAttribute
\indicator_{\iImage, \iAttribute}
|\attribute_{\iImage,\iAttribute} - \attribute^\gt_{\iImage,\iAttribute}|^2
.
\end{equation}

For the mask few-shot supervision, \
we employ the volume rendering in~\eq{volren} to project the 3D volumetric neural mask field $\mask_\iAttribute(\bx)$ into image space, and then supervise it as:
\newcommand{\crossentropy}[2]{\text{CE}\left(#1,#2\right)}
\begin{equation}
\loss{mask}(\cdot) = \!\!\sum_{\iImage, \iAttribute}
\indicator_{\iImage, \iAttribute} \:
\expect_\ray
\left[ 
\crossentropy
{\Mask(\ray \given \latent_\iImage, \pars)}
{\Mask^\gt_{\iImage, \iAttribute}(\ray)}
\right]
,
\label{eq:mask}
\end{equation}
where $\crossentropy{\cdot}{\cdot}$ denotes cross entropy, and the field $\sigma(\bx)$ in \eq{volren} is learned by minimizing \eq{recon}.
Importantly, as we do not wish for \eq{mask} to interfere with the training of the underlying 3D representation learned through \eq{recon}, we \textit{stop gradients} in \eq{mask} w.r.t. $\sigma(\bx)$.
Furthermore, in practice, because the attribute mask vs. background distribution can be highly imbalanced depending on which attribute the user is trying to control (\eg an eye only covers a very small portion of an image), we employ a \textit{focal loss}~\cite{lin2017focal} in place of the standard cross entropy loss.

\subsection{Controlling and rendering images}
\label{sec:inference}
In what follows, we drop the image subscript $\iImage$ to simplify notation without any loss of generality.
Given a latent code $\latent$ representing the 3D scene behind an image, we derive a mapping to our attributes via a neural map $\Attribute$ with learnable parameters $\pars$:
\begin{equation}
\{ \attribute_\iAttribute \} = \AttributeNet(\latent \given \pars), \quad 
\AttributeNet : \real^\latentDim \rightarrow [0,1]^\nAttributes
,
\end{equation}
where these correspond to the \textit{sliders} in \Figure{teaser}.
In the same spirit of \eq{hypermap}, to allow for complex topological changes that may not be represented by the change in a single scalar value alone, we lift the attributes to a hyperspace.
In addition, since each attribute governs different aspects of the scene, we employ \textit{per-attribute} learnable hypermaps $\{\hypermap_\iAttribute\}$, which we write:
\begin{align}
\attribute_\iAttribute(\point) &= \hypermap_\iAttribute(\point, \attribute_\iAttribute \given \pars) \quad \hypermap_\iAttribute: \real^3 \times \real \rightarrow \real^\hyperAttributeDim
.
\end{align}
Note that while $\attribute_\iAttribute$ is a scalar \textit{value}, $\attribute_\iAttribute(\point)$ is a \textit{field} that can be queried at any point $\point$ in space.
These fields are concatenated to form $\attributes(\point)=\{\attribute_\iAttribute(\point)\}$.
 
We then provide all this information to generate an \textit{attribute masking field} via a network $\MaskNet(\cdot \given \pars)$.
This field determines which attribute \textit{attends} to which position in space~$\point$:
\begin{align}
\label{eq:mask_c}
\mask_0(\point)& \oplus \mask(\point) = \MaskNet(\Canonicalizer(\point), \bbeta(\point), \attributes(\point) \given \pars), \\
\MaskNet &: \real^3 \times \real^\latentDim \times \real^{\nAttributes \times \hyperAttributeDim} \rightarrow \real^{\nAttributes+1}_+
,
\end{align}
where $\oplus$ is a concatenation operator, $\mask(\point){=}\{\mask_\iAttribute(\point)\}$, and the additional mask $\mask_0(\point)$ denotes space that is not affected by \textit{any} attribute.
Note that because the mask location should be affected by both the particular attribute of interest (\eg, the selected eye status) and the global appearance of the scene (\eg, head movement),
$\MaskNet$ takes both $\bbeta(\point)$ and $\attributes(\point)$ as input in addition to $\Canonicalizer(\point)$.
In addition, because the mask is modeling the attention related to attributes, collectively, these masks satisfy the partition of unity property:
\begin{align}
\mask_0(\point) + \Sigma_\iAttribute [\mask_\iAttribute(\point)] = 1 \quad \forall \point \in \real^3
.
\end{align}
Finally, in a similar spirit to~\eq{hypernerf}, all of this information is processed by a neural network that produces the desired radiance and density fields used in volume rendering:
\begin{equation}
\begin{rcases}
\bc(\point) \\
\sigma(\point)
\end{rcases} \!\!=\! \Representation(\Canonicalizer(\point),
\underbrace{\mask(\point) \odot \attributes(\point)}_\text{attribute controls},
\underbrace{\mask_0(\point) \cdot \latent(\point)}_\text{everything else}
\given \pars)
.
\label{eq:representation}
\end{equation}
In particular, note that $\mask(\point){=}0$ implies $\mask_0(\point){=}1$, hence our solution has the capability of reverting to classical HyperNeRF~\eq{hypernerf}, where all change in the scene is globally encoded in $\bbeta(\point)$.
Finally, these fields can be used to render the mask in image space, following a process analogous to volume rendering of radiance:
\begin{equation}
\Mask(\ray\!\given\!\allpars) \!=\!\!
\int_{t_n}^{t_f} \!\!\!\! T(t) \cdot \sigma(\br(t))  \cdot
[\mask_0(\ray(t)) \oplus \mask(\ray(t))] \, dt
.
\label{eq:volren}
\end{equation}
We depict our inference flow in \Figure{pipeline}~(b).

\subsection{Implementation details}
We implement our method for NeRF based on the JAX \cite{jax2018github} implementation of HyperNeRF~\cite{park2021hypernerf}.
We use both the scheduled windowed positional encoding and weight initialization of \cite{park2020deformable}, as well as the coarse-to-fine training strategy~\cite{park2021hypernerf}.

Besides the newly added networks, we follow the same architecture as HyperNeRF.
For the attribute network $\Attribute$ we use a six-layer multi-layer perceptron (MLP) with 32 neurons at each layer, with a skip connection at the fifth layer, following \cite{park2020deformable,park2021hypernerf}.
For the lifting network $\hypermap_\iAttribute$, we use the same architecture as $\hypermap$, except for the input and output dimension sizes.
For the masking network $\MaskNet$ we use a four-layer MLP with 128 neurons at each layer, followed by an additional 64 neuron layer with a skip connection.
The network $\Representation$ also shares the same architecture as HyperNeRF, but with a different input dimension size to accommodate for the changes our method introduces.

\paragraph{2D implementation}
To show that our idea is not limited to neural radiance fields, we also test a 2D version of our framework that can be used to directly represent images, without going through volume rendering.
We use the same architecture and training procedure as in the NeRF case, with the exception that we do not predict the density $\sigma$, and we also do not have the notion of depth---each ray is directly the pixel.
We center crop each video and resize each frame to be $128\times128$.

\paragraph{Hyperparameters}
We train all our NeRF models with $480\times270$ images and with 128 samples per ray.
We train for 250k iterations with a batch size of 512 rays.
During training, we also maintain that 10\% of rays are sampled from annotated images.
We set $\loss{attr}=10^{-1}$, $\loss{mask}=10^{-2}$ and $\loss{enc}=10^{-4}$.
For the number of hyper dimensions we set $\hyperAttributeDim = 8$.
For the 
experiments with the 2D implementation, we sample 64 random images from the scene and further subsample 1024 pixels from each of them.
For all experiments we use Adam~\cite{kingma2014adam} with learning rate $10^{-4}$ exponentially decaying to $10^{-5}$, as it reaches 250k iterations.
We provide additional architectural details in the supplementary material. 
Training a single model takes around 12 hours on an NVIDIA V100 GPU.
 
\begin{figure*}
    \centering
    \includegraphics[width=\linewidth]{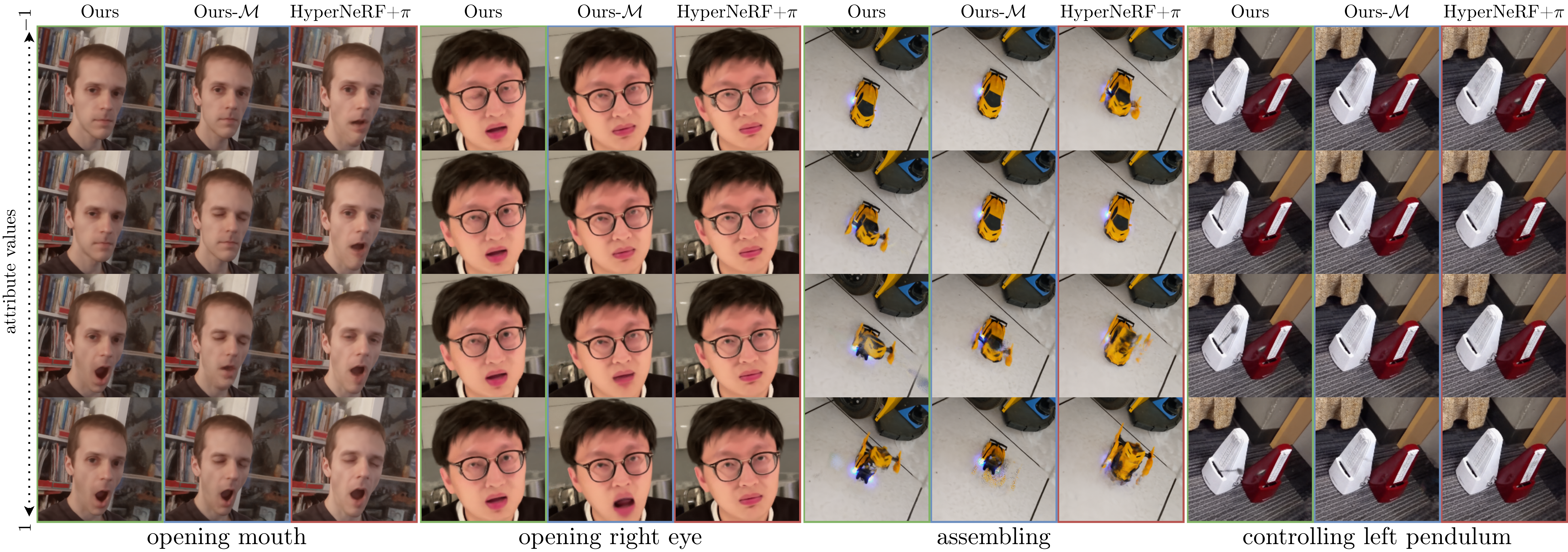}
    \caption{{\bf Novel view and novel attribute synthesis on real data --}
    We synthesize scenes from a novel view and with a novel attribute combination, not seen during training.
    A naive extension of HyperNeRF, HyperNeRF{+}$\pi$ fails to disentangle attributes and results in a modification of the scene irrespectively of attribute meaning %
    \eg, opening mouth results in closing eyes at the same time. %
    Ours-$\MaskNet$ improves the results, but does not disentangles the attribute space, as successfully done by our complete method. %
    The differences between these methods can even lead to complete failure cases, as shown in the metronome and the toy car case.
    }
    \label{fig:novel_view}
\end{figure*}

\section{Results}
\label{sec:results}

\subsection{Datasets and baselines}

We evaluate our method on two datasets: real video sequences captured with a smartphone ({\it real dataset}) and synthetically rendered sequences ({\it synthetic dataset}). Here we introduce those datasets and the baselines for our approach.

\paragraph{Real dataset}
Each of the seven real sequences is 1 minute long and was captured either with a Google Pixel 3a or an Apple iPhone 13 Pro.
Four of them consists of people performing different facial expressions including smiling, frowning, closing or opening eyes, and opening mouth.
For the other three, we captured a toy car changing its shape ({\it a.k.a.} Transformer), a single metronome, and two metronomes beating with different rates.
For one of the four videos depicting people, to use it for the 2D implementation case, we captured it with a static camera that shows a frontal view of the person.
All other sequences feature camera motions showing front and sides of the object in the center of the scene.
For videos with human subjects, the subjects signed a participant consent form, which was approved by a research ethics board.
We informed the participants that their data will be altered with our method.

We extract frames at 15 FPS which gives approximately 900 frames per capture. 
Because novel attribute synthesis via user control on real scenes does not have a ground truth view---we aim to create scenes with unseen attribute combinations---the benefit of our method is best seen qualitatively.
Nonetheless, to quantitatively evaluate the rendering quality, we interpolate between two frames and evaluate its quality.
In more detail, to minimize the chance of the dynamic nature of the scene interfering with this assessment, we use every other frame as a test frame for the interpolation task.

For all human videos, we define three attributes---one for the status of each of the two eyes, and one for the mouth.
We annotate only six frames per video in this case, specifically the frames that contain the extremes of each attribute (\eg, left eye fully open).
For the toy car, we set the shape of the toy car to be an attribute, and annotate two extremes from two different view points---when the toy is in robot-mode and when it is in car-mode from its left and right side.
For the metronomes, we consider the state of the pendulum to be the attribute
and annotate the two frames with the two extremes for the single metronome case, and seven frames for the two metronome case as the pendulums of the two metronomes are often close to each other and required special annotations for these close-up cases; see~\Figure{novel_view}.

\paragraph{Synthetic dataset} Since the lack of ground-truth data renders measuring the quality of novel attribute synthesis infeasible in practice, %
we leverage Kubric software~\cite{kubric2021github} to generate synthetic dataset, where we know exactly the state of each object in the scene. 
We create a simple scene where three 3D objects, the teapot \cite{teapot}, the Stanford bunny \cite{bunny}, and Suzanne \cite{suzanne}, are placed within the scene and are rendered with varying surface colors, which are our attributes; see \Figure{kubric3d}.
We generate 900 frames for training and 900 frames for testing.
To ensure that the attribute combination during training is not seen in the test scene, we set the attributes to be synchronized for the training split, and desynchronized for the test split. 
We further render the test split from different camera positions than the training split to account for novel views.
We randomly sample 5\% of the frames with a given attribute for each object to be set as the ground-truth attribute. During validation, we use attribute values directly to predict the image.

\paragraph{Baselines}
To evaluate the reconstruction quality of our method, CoNeRF, we compare it with four different baselines: 
\CIRCLE{1}~standard NeRF~\cite{mildenhall2020nerf};
\CIRCLE{2}~NeRF+Latent, a simple extension to NeRF where we concatenate each coordinate $\bx$ with a learnable latent code $\bbeta$ to support appearance changes of the scene;
\CIRCLE{3}~Nerfies \cite{park2020deformable};
and \CIRCLE{4}~HyperNeRF\footnote{We use the version with dynamic plane slicing as it consistently outperforms the axis-aligned strategy; see \cite{park2021hypernerf} for more details.}~\cite{park2021hypernerf}.
Additionally, as existing methods do not support attribute-based control with a few-shot supervision, 
we create another baseline \CIRCLE{5} by extending HyperNeRF with a simple linear regressor $\pi$ that regresses $\bbeta_c$ given $\balpha_c$.
We call this baseline HyperNeRF{+}$\pi$.
To further show the importance of masking, we also compare our approach against a stripped-down version of our method,
Ours{-}$\MaskNet $,
where we disable the part of our pipeline responsible for masking.
All baselines that utilize annotations were trained with the same sparse labels as our method.

\subsection{Comparison with the baselines}
\begin{figure}
    \centering
    \begin{subfigure}[b]{0.164\linewidth}
        \centering
        \includegraphics[width=\linewidth]{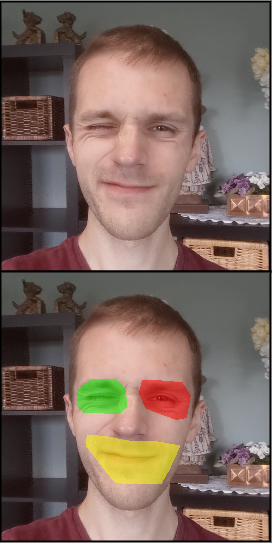}
        \caption{annotation}
    \end{subfigure}
    \hfill{}
    \begin{subfigure}[b]{0.816\linewidth}
        \centering
        \includegraphics[width=\linewidth]{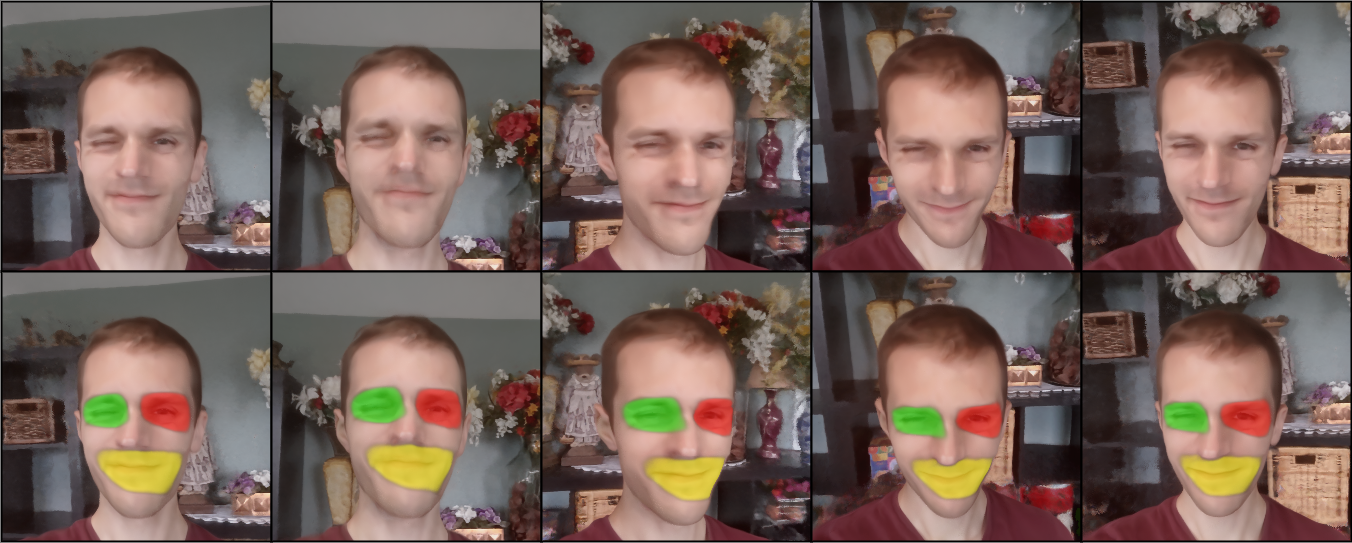}
        \caption{unannotated views}
    \end{subfigure}
    \caption{{\bf Annotation example -- }
    We provide only a rough annotation for each attribute, which is enough for the method to discover the mask for each attribute across all views automatically.
    Bottom row shows masks overlaid on the image.
    }
    \label{fig:annotate}
\end{figure}

\vspace{-\customparskip}
\paragraph{Qualitative highlights}
We first show qualitative examples of novel attribute and view synthesis on the real dataset in \Figure{novel_view}.
Our method allows for controlling the selected attribute without changing other aspects of the image---our control is disentangled.
This disentanglement allows our method to generate images with attribute combinations that were not seen at training time.
On the contrary, as there is no incentive for the learned embeddings of HyperNeRF to be disentangled, the simple regression strategy of HyperNeRF{+}$\pi$ results in entangled control, where when one tries to close/open the mouth it ends up affecting the eyes.
The same phenomenon happens also for Ours{-}$\MaskNet$.
Moreover, due to the complexity of motions in the scene, HyperNeRF{+}$\pi$ fails completely to render novel views of the toy car, whereas our method, with only four annotated frames, successfully provides both controllability and high-quality renderings.
Please also see \SupplementaryMaterial for more qualitative results, including a video demonstration.

Note that in all of these sequences, we provide highly sparse annotations and yet our method learns how each attribute should influence the appearance of the scene.
In \Figure{annotate}, we show an example annotation and how the method finds the mask for unannotated views. 

\begin{table}
\centering
\begin{tabular}{@{}lccc@{}}
\toprule
Method & PSNR$\uparrow$ & MS-SSIM$\uparrow$ & LPIPS$\downarrow$ \\
\midrule
  HyperNeRF{+}$\pi$ & 25.963 &   0.854 &  0.158 \\
                   \textbf{Ours}{-}$\MaskNet$ & 27.868 &   0.898 &  0.155 \\
                                \textbf{Ours} & \textbf{32.394} &   \textbf{0.972} &  \textbf{0.139} \\
\bottomrule
\end{tabular}
\caption{
\textbf{Novel view and novel attributes results -- } 
We report average PSNR, MS-SSIM, and LPIPS values for novel view and novel attribute synthesis on synthetic data. 
Our method gives the best results.
} %
\label{tab:kubric3d}
\end{table}
\begin{figure}
    \centering
    \includegraphics[width=\linewidth]{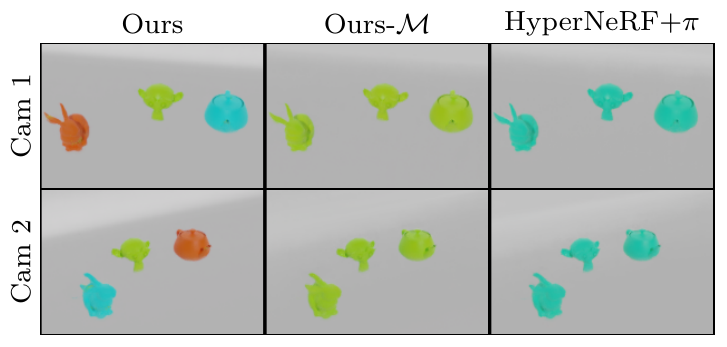}
    \caption{{\bf Novel view and novel attribute synthesis on synthetic data -- }
    We show examples of novel view and novel attribute synthesis on synthetic data.
    The scene is composed of three objects, where the color of each object is their attribute.
    Our method provides control over the color of each object independently, whereas both HyperNeRF{+}$\pi$ and Ours{-}$\MaskNet$ fail to deliver controllability and results in all three objects having the same attribute in the rendered scene.
    }
    \label{fig:kubric3d}
    \vspace{-1.0em}
\end{figure}
\paragraph{Quantitative results on synthetic dataset}
To complete the qualitative evaluation of our method,
we provide results using synthetic dataset with available ground truth.
We measure Peak Signal-to-Noise Ratio (PSNR), Multi-scale Structural Similarity (MS-SSIM)~\cite{wang2003multiscale}, and Learned Perceptual Image Patch Similarity (LPIPS)~\cite{zhang2018unreasonable} and report them in \Table{kubric3d}.
With only 5\% of the annotations, our method provides the best novel-view and novel-attribute synthesis results, as reconfirmed by the qualitative examples in \Figure{kubric3d}.
As shown, neither HyperNeRF{+}$\pi$ nor Ours{-}$\MaskNet$ is able to provide good results in this case, as without disentangled control of each attribute, the novel attribute and view settings of each test frame cannot be synthesized properly.

\paragraph{Interpolation task}
To further verify that our rendering quality does not degrade with the introduction of controllability, 
we evaluate our method on a frame interpolation task without any attribute control. 
Unsurprisingly, as shown in \Table{sota}, all methods that support dynamic scenes work similarly, including ours for interpolation.
Note that for the interpolation task, we interpolate every other frame, in order to minimize the chance of attributes affecting the evaluation. 
Here, we are purely interested in the rendering quality from a novel view.

\begin{table}
\centering
\resizebox{\linewidth}{!}{ %
\begin{tabular}{@{}lccc@{}}
\toprule
Method &  PSNR $\uparrow$ & MS-SSIM $\uparrow$ & LPIPS $\downarrow$  \\
\midrule
                                   NeRF & 28.795 &   0.951 &  0.210 \\
NeRF + Latent \cite{mildenhall2020nerf} & 32.653 &   0.981 &  0.182 \\
      NeRFies \cite{park2020deformable} & 32.274 &   0.981 &  0.180 \\
  HyperNeRF \cite{park2021hypernerf} & 32.520 &   0.981 &  0.169 \\
\midrule
\textbf{Ours}{-}$\MaskNet$ & 32.061 &   0.979 &  0.167 \\
\textbf{Ours} & 32.342 &   0.981 &  0.168 \\

\bottomrule
\end{tabular}
}
\caption{
\textbf{Quantitative results (interpolation) -- }
We report results in terms of PSNR, MS-SSIM, and LPIPS for the interpolation task. 
These results are obtained for interpolated view synthesis only, not for novel attribute rendering.
Our method provides similar performance in terms of rendering quality, but with controllability.
} %
\label{tab:sota}
\end{table}

\begin{figure}
    \centering
    \includegraphics[width=\linewidth, trim=0 320 0 0, clip]{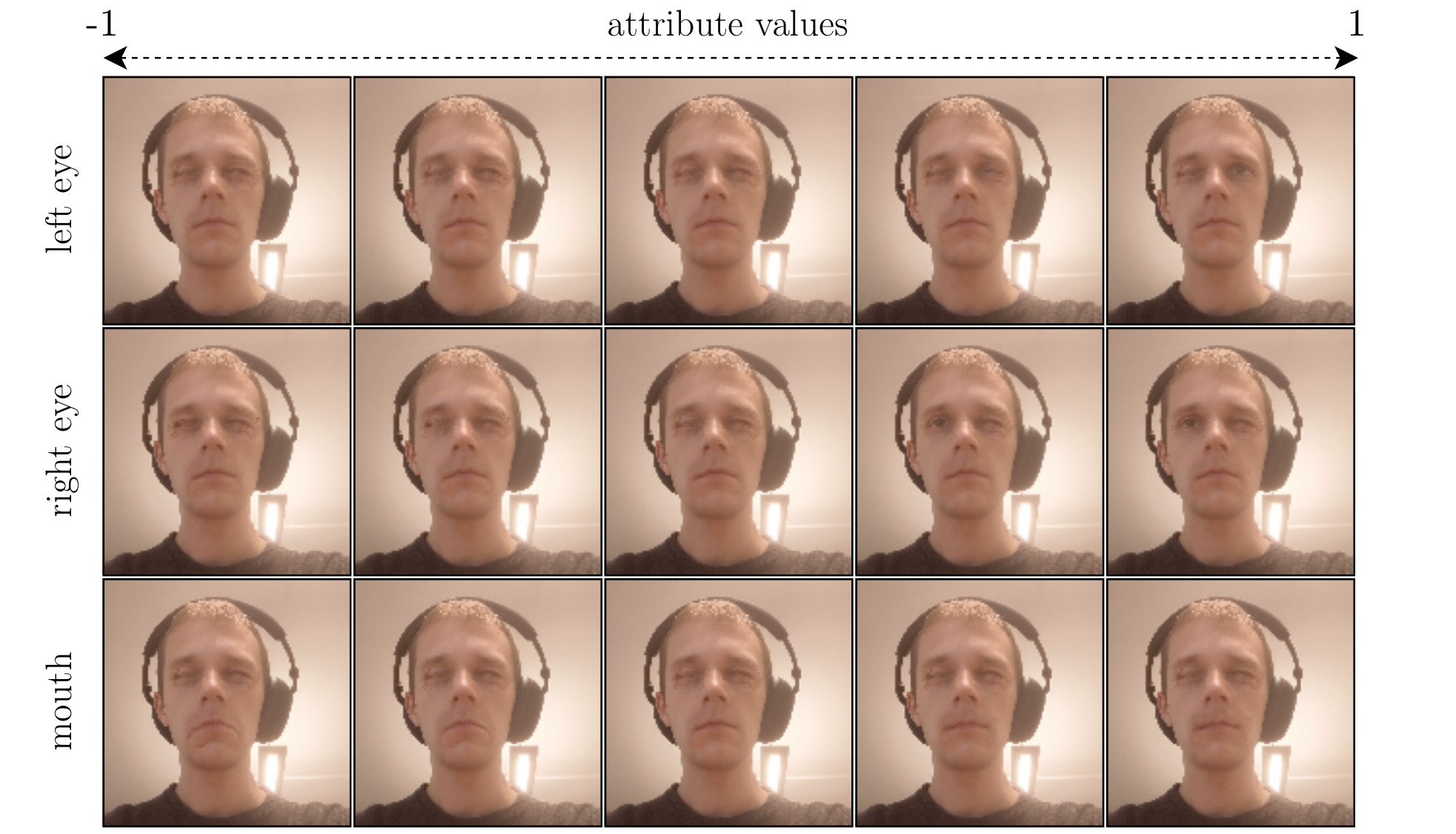}
    \caption{
    {\bf 2D image generation example -- } 
    Our framework also generalizes to direct generation of 2D images.
    Here we show novel attribute synthesis for a webcam video of a person making expressions.
    Each individual part of the scene is correctly controlled according to the attribute values.}
    \label{fig:2d-interpolation}
\end{figure}
\subsection{Direct 2D rendering}
To verify how our approach generalizes beyond NeRF models and volume rendering, we apply our method to videos taken from a single view point, creating a 2D rendering task. 
We show in \Figure{2d-interpolation} a proof-of-concept for employing our approach outside of NeRF applications to allow controllable neural generative models.

\subsection{Ablation study}
\begin{table}
\centering
\normalsize
\resizebox{\linewidth}{!}{ %
\setlength{\tabcolsep}{2pt}
\begin{tabular}{@{}lccc ccc@{}}
\toprule
& \multicolumn{3}{c}{Real (interpolation)} &  \multicolumn{3}{c}{Synthetic (novel view \& attr.)}  \\
\cmidrule(lr){2-4}
\cmidrule(lr){5-7}
Model & PSNR $\uparrow$ & MS-SSIM $\uparrow$ & LPIPS $\downarrow$ & PSNR $\uparrow$ & MS-SSIM $\uparrow$ & LPIPS $\downarrow$ \\
\midrule
                    Base ($\loss{recon}$) & 32.457 &   0.981 &  0.168 &  24.407 &   0.718 &  0.173 \\
                            $+\loss{enc}$ & 32.478 &   0.982 &  0.167 & 27.018 &   0.871 &  0.164 \\
              $+ \loss{enc} +\loss{attr}$ & 32.254 &   0.981 &  0.167 & 27.322 &   0.873 &  0.147 \\
$+ \loss{enc} + \loss{attr} +\loss{mask}$ & 32.342 &   0.981 &  0.168 & \textbf{32.394} &  \textbf{0.972} &  \textbf{0.139}\\
\bottomrule
\end{tabular}
} %
\caption{
\textbf{{Effect of loss functions --}} 
We report the rendering quality of our method as we procedurally introduce the loss terms.
For controlled rendering with novel views and attributes (synthetic data), each loss term adds to the rendering quality, with the $\loss{mask}$ being critical.
For the novel view rendering on real data, addition of loss functions for controllability do not have a significant effect on the rendering quality---they do no harm.
} %
\label{tab:ablations}
\end{table}
\vspace{-\customparskip}
\paragraph{Loss functions}
In \Table{ablations}, we show how each loss term affects the network's performance, contributing to performance improvements.
When rendering novel views with novel attributes, the full formulation is a must, as without all loss terms the performance drops significantly---for example, results without $\loss{mask}$ is similar to Ours-$\MaskNet$ results in \Table{kubric3d} and \Figure{kubric3d}.
In the case of the interpolation task, the additional loss functions for controllability have no significant effect on the rendering quality.
In other words, our controllability losses \textbf{do not interfere} with the rendering quality, other than imbuing the framework with controllability.\vspace{-0.2em}

\begin{figure}
    \centering
    \includegraphics[width=\linewidth]{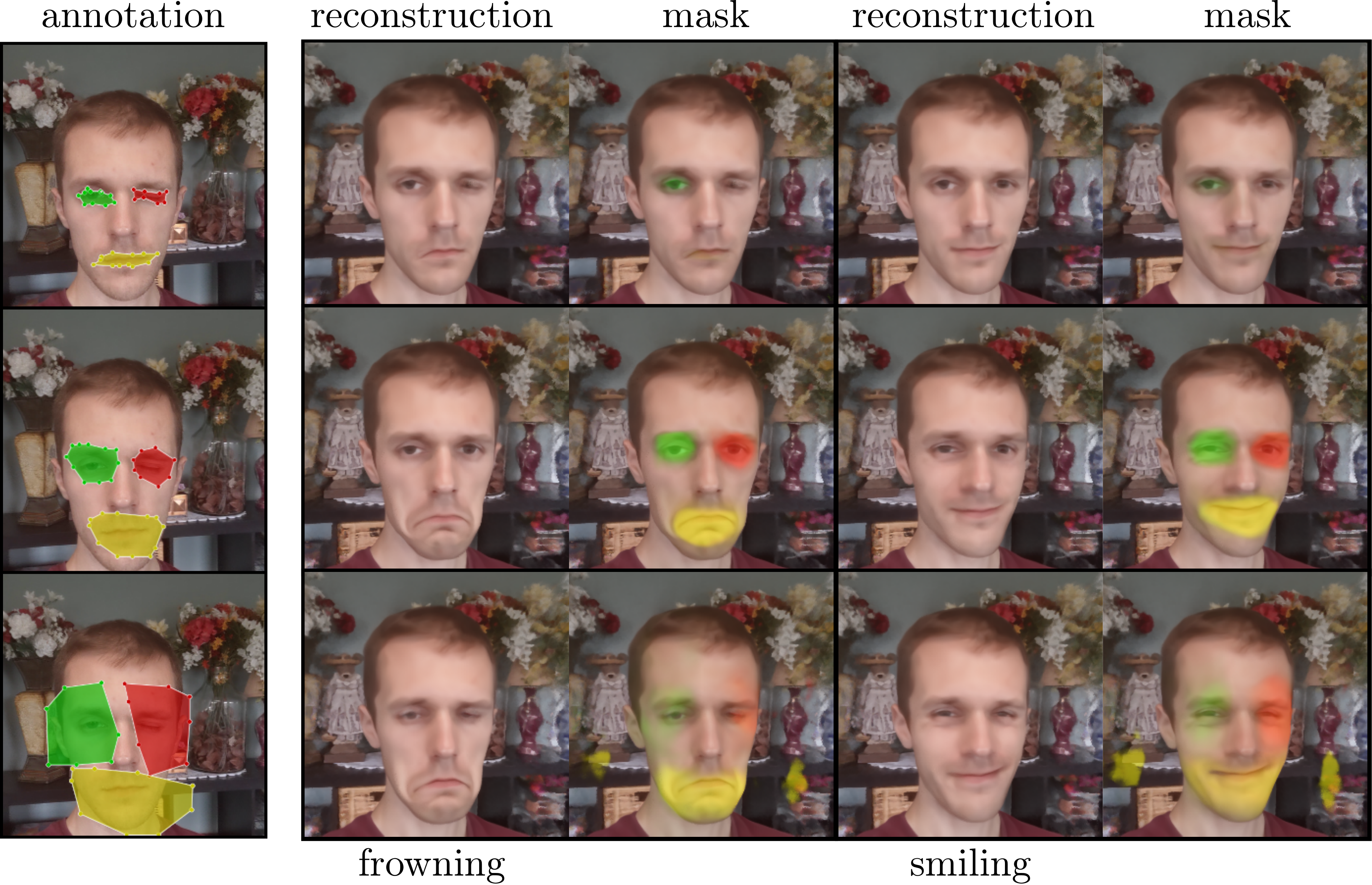}
    \vspace{-1.8em}
    \caption{{\bf Effect of annotation quality -- }
    Our method is moderately robust to the quality of annotations. 
    We visualize the results for two expressions: frowning and smiling, while keeping both eyes in a neutral position.
    Even with wildly varying annotations as shown, the reconstructions are reasonably controlled, with the exception of the top row, where we show a case where the annotations is too restrictive, resulting in the annotation being ignored for one eye.
    We show in bottom row also an interesting case, where the mask is large enough to start capturing the correlation among mouth expressions and the eye.
    }
    \vspace{-1.2em}
    \label{fig:ablation_annotate}
\end{figure}
\paragraph{Quality of few shot supervision}
We test how sensitive our method is against the quality of annotation supervision.
In \Figure{ablation_annotate} we demonstrate how each annotation leads to the final rendering quality.
Our framework is robust to a moderate degree to the inaccuracies in the annotations.
However, when they are too restrictive, the mask may collapse, as shown on the top row.
Too large of a mask could also lead to moderate entanglement of attributes, as shown in the bottom row.
Still, in all cases, our method provides a reasonable control over what is annotated.\vspace{-0.2em}

\begin{figure}
    \centering
    \begin{subfigure}[t]{0.38\linewidth}
        \centering
        \includegraphics[width=\linewidth]{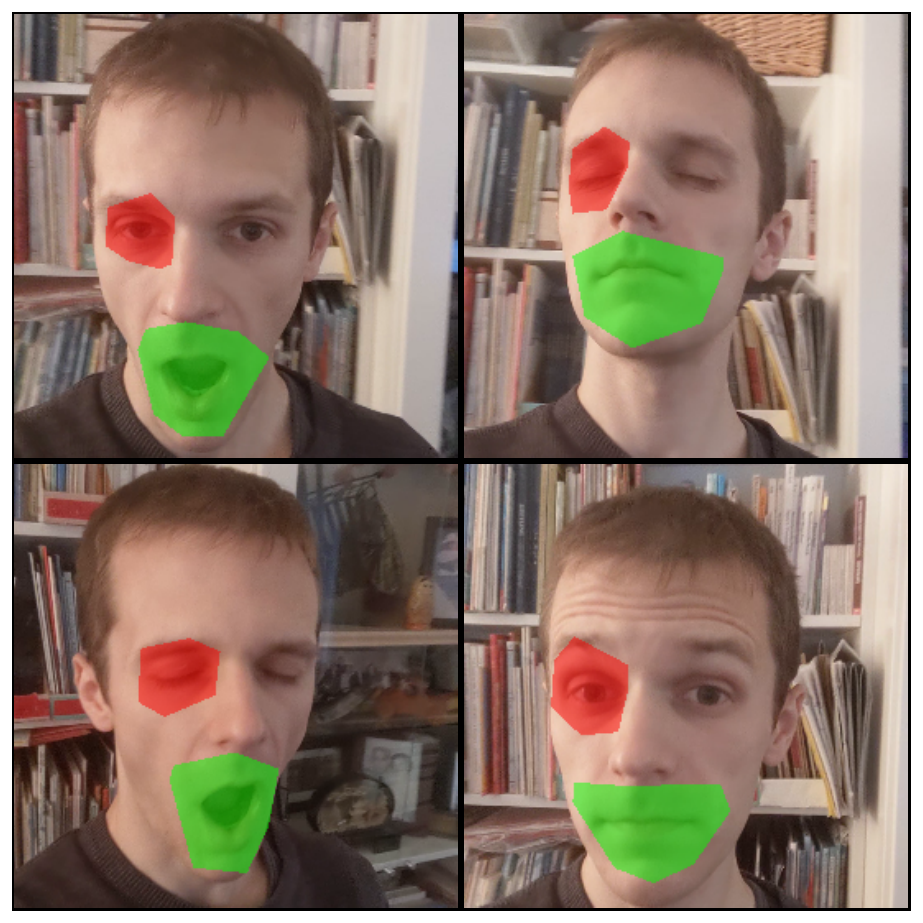}
        \caption{annotated samples}
    \end{subfigure}
    \hfill{}
    \begin{subfigure}[t]{0.595\linewidth}
        \centering
        \includegraphics[width=\linewidth]{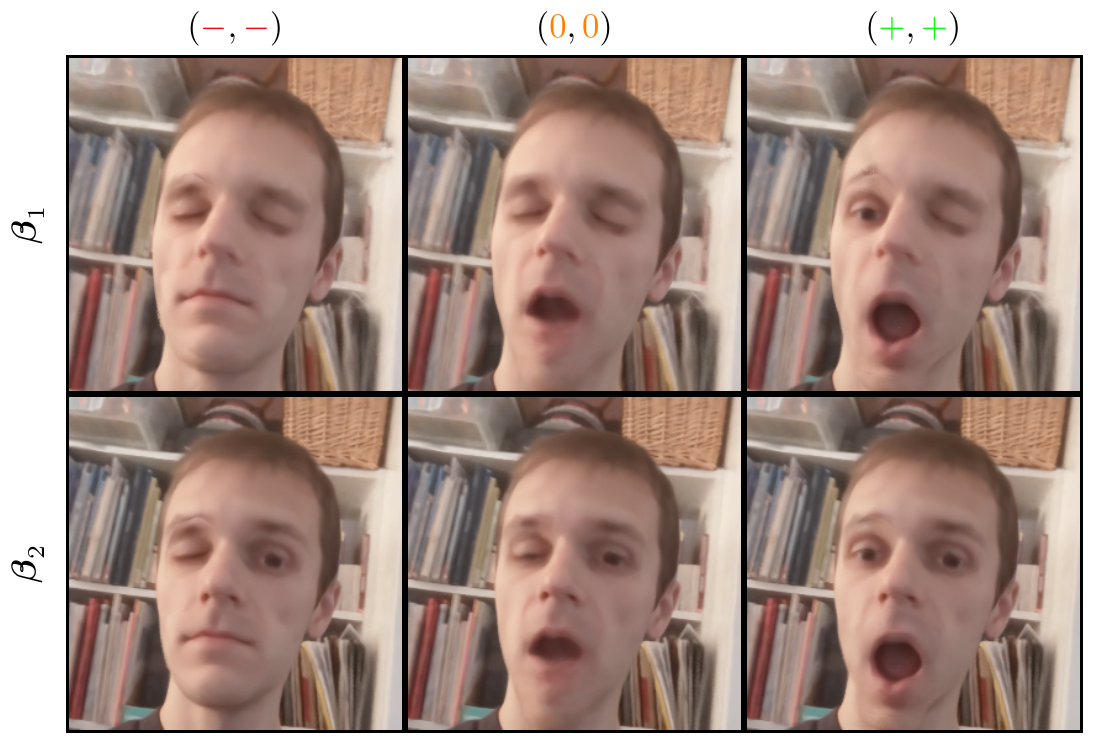}
        \caption{renderings}
    \end{subfigure}
    \caption{{\bf Example with unannotated attributes --}
    We show an example of how our method performs when a part of the image changes appearance, but is not annotated.
    With the annotations in (a), we synthesize the scene with novel view and attributes in (b), where the two rows are with different $\bbeta$ configurations. 
    We denote the attribute configuration on the top of each column in (b).
    As shown, the change that is not annotated is simply encoded in the per-image encoding $\bbeta$.
    }
    \label{fig:ablation_unannotated}
\end{figure}
\paragraph{Unannotated attributes}
A natural question to ask is then what happens with the unannotated changes that may exist in the scene.
In \Figure{ablation_unannotated} we show how the method performs when annotating only parts of the appearance change within the scene. 
The unannotated changes of the scene get encoded as $\bbeta$, as in the case of HyperNeRF~\cite{park2021hypernerf}.

\section{Conclusions}
We have introduced CoNeRF, an intuitive controllable NeRF model that can be trained with few-shot annotations in the form of attribute masks.
The core contribution of our method is that we represent attributes as localized masks, which are then treated as latent variables within the framework.
To do so we regress the attribute and their corresponding masks with neural networks.
This leads to a few-shot learning setup, where the network learns to regress provided annotations, and if they are not provided for a given image, proper attributes and masked are discovered throughout training automatically.
We have shown that our method allows users to easily annotate what to control and how, within a single video simply by annotating a few frames, which then allows rendering of the scene from novel views and with novel attributes, at high quality.

\paragraph{Limitations}
While our method delivers controllability to NeRF models, there is room for improvement.
First, our disentanglement of attribute strictly relies on the locality assumption---if multiple attributes act on a single pixel, our method is likely to have entangled outcomes when rendering with different attributes. 
An interesting direction would therefore be to incorporate manifold disentanglement approaches~\cite{li2020markov,zhang2021product} to our method.
Second, while very few, we still require sparse annotations.
Unsupervised discovery of controllable attributes, for example as in \cite{kulkarni2019unsupervised}, in a scene remains yet to be explored.
Lastly, we resort to user intuition on which frames should be annotated---we heuristically choose frames with extreme attributes (\eg, mouth fully open).
While this is a valid strategy, an interesting direction for future research would be to employ active learning techniques for this purpose~\cite{ren2020survey,belharbi2021deep}

We further discuss potential societal impact of our work in the \SupplementaryMaterial.

\iftoggle{cvprfinal}{%
\section{Acknowledgements}
\label{sec:ack}
We thank  Thabo Beeler, JP Lewis, and Mark J. Matthews for their fruitful discussions, and Daniel Rebain for helping with processing the synthetic dataset. The work was partly supported by National Sciences and Engineering Research Council of Canada (NSERC), Compute Canada, and Microsoft Mixed Reality \& AI Lab.

}{}

{
    \small
    \bibliographystyle{ieee_fullname}
    \bibliography{macros,main}
}

\appendix

\setcounter{page}{1}

\twocolumn[
\centering
\Large
\textbf{CoNeRF: Controllable Neural Radiance Fields} \\
\vspace{0.5em}Supplementary Material \\
\vspace{1.0em}
] %
\appendix

\section{Potential social impact}
\ethicspart{
Our work is originally intended for creative and entertainment purposes, for example to allow users to easily edit their personal photos to have all the members of a group photo to have their eyes open.
However, as with all work that enable editable models, our method has the potential to be misused for malicious purposes such as deep fakes.
We strongly advise against such misuse.
Recent work~\cite{asnani2021reverse} has shown that it is possible to detect deep fakes, hinting that it should be possible to detect these deep learning-generated images.
One of our future research direction is also along these lines, where we now aim to reliably detect images generated by our method.
}

\section{Architecture details}
\begin{figure}
    \centering
    \includegraphics[width=\linewidth]{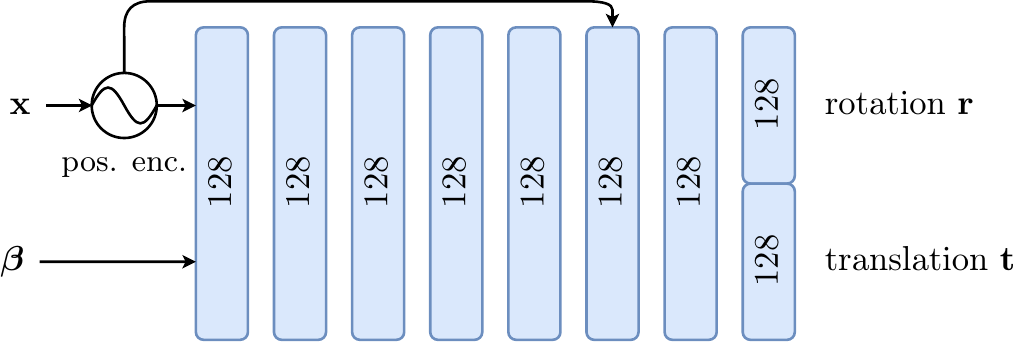}
    \caption{The canonicalization network takes positionally encoded raw coordinates $\bx$ and learnable per-image latent code $\bbeta$ and outputs rotation $\mathbf{r}$ expressed as a quaternion and translation $\mathbf{t}$. We rigidly transform each point $\bx$ with an affine transform using both output. We use windowed positional encoding \cite{park2020deformable} for $\bx$ with 8 components, linearly increasing contribution of components throughout 80k steps. We initialize the last layer to small values so the network can learn a base structure of the data.}
    \label{fig:warping-field}
\end{figure}

\begin{figure}
    \centering
    \includegraphics[width=0.9\linewidth]{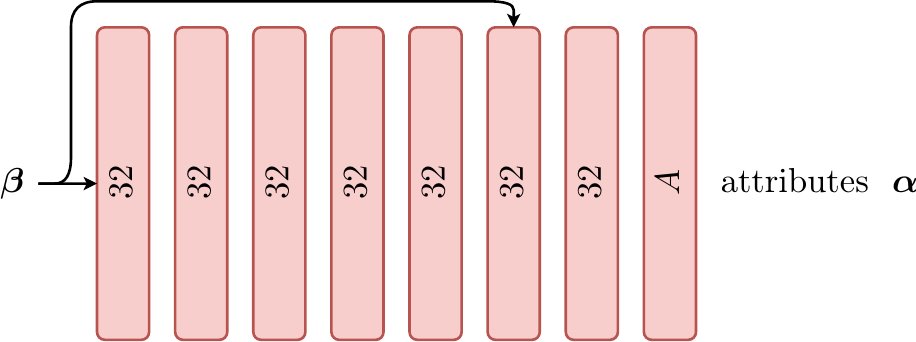}
    \caption{The attribute map $\AttributeNet$ takes a per-image learnable latent code $\bbeta$ and outputs $A$ attributes $\attributes$.}
    \label{fig:attribute-net}
\end{figure}

\begin{figure}
    \centering
    \includegraphics[width=\linewidth]{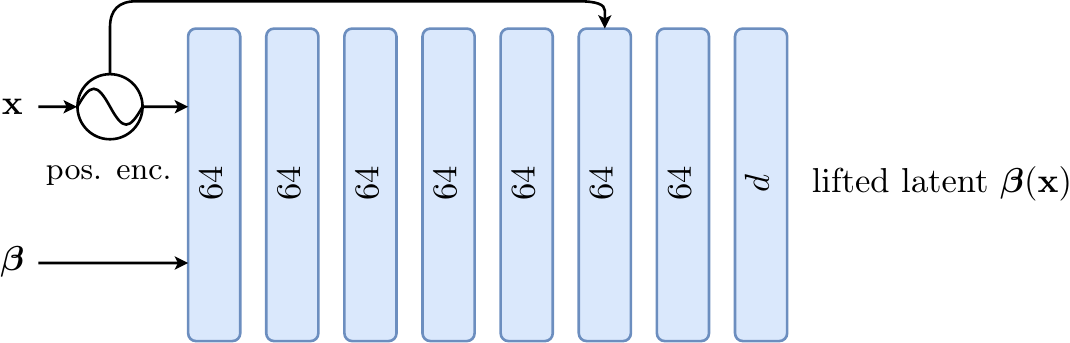}
    \caption{The network predicting lifted latent code $\bbeta$, takes per-image $\bbeta$ as an input, positionally encoded raw points $\bbeta$ and outputs a lifted code of size $d$. We use only one sine component to encode $\bx$.}
    \label{fig:hypermap}
\end{figure}

\begin{figure}
    \centering
    \includegraphics[width=\linewidth]{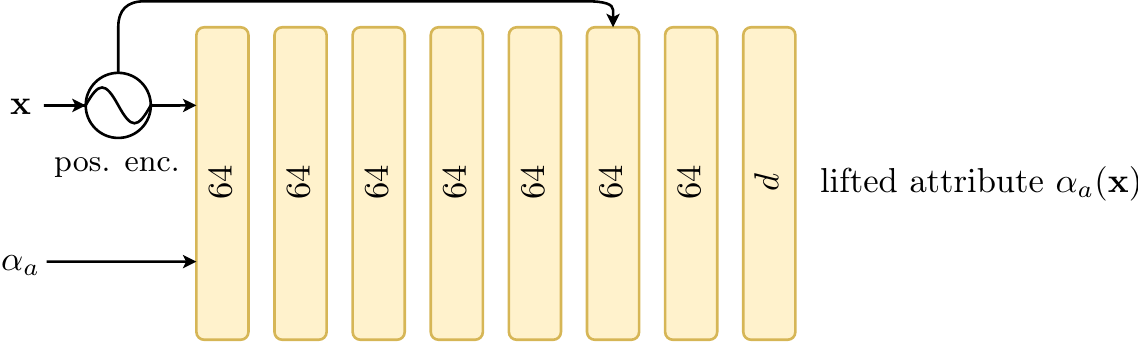}
    \caption{Per-attributes hypermaps take an attribute together with encoded $\bx$ coordinates and output lifted $\attribute_a(\bx)$ ambient code of size $d$. We encode $\bx$ with only single component.}
    \label{fig:hypermap-alpha}
\end{figure}

\begin{figure}
    \centering
    \includegraphics[width=0.7\linewidth]{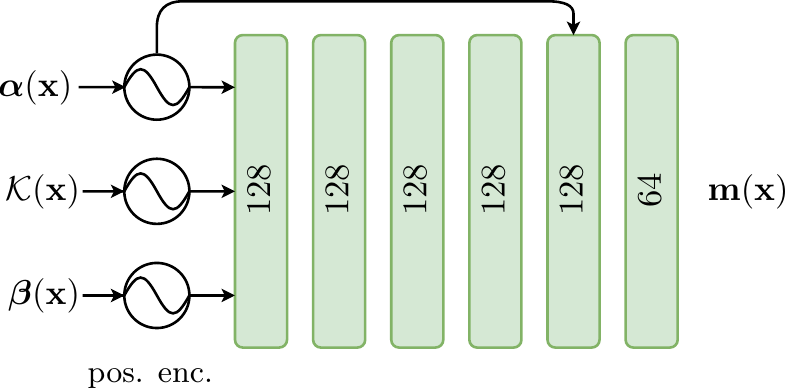}
    \caption{Masking network $\MaskNet$ take lifted attributes $\attributes(\bx)$, lifted latent code $\bbeta(\bx)$ and canonicalized points $\Canonicalizer(\bx)$. We transform $\attributes(\bx)$ and $\bbeta(\bx)$ through a windowed positional encoding where we start at 1k-th step linearly increasing a single sine component for the next 10k steps. Points $\Canonicalizer(\bx)$ are encoded with 8 components. The output is activated with a sigmoid function. We realize $\mathbf{m}_0(\bx)$ as $\mathbf{m}(\bx)_0 = 1 - \sum_{a \in A} \mathbf{m}_a(\bx)$, and clip the output to ensure the values range to be in $[0,1]$. Note that while the network shares similarities with the radiance field prediction part $\Representation$, it is not conditioned on view directions and appearance codes.
    } 
    \label{fig:masknet}
\end{figure}

\begin{figure}
    \centering
    \includegraphics[width=\linewidth]{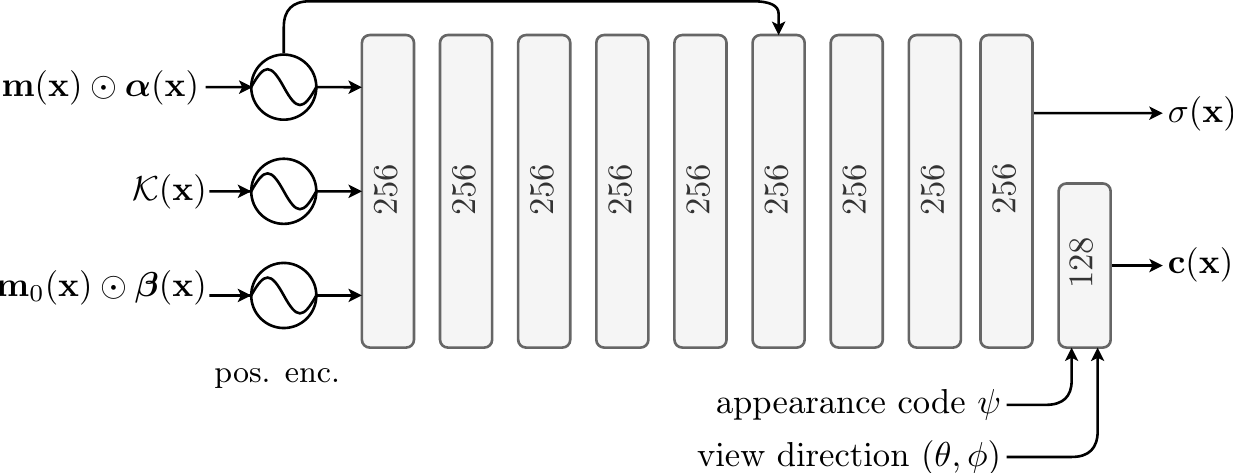}
    \caption{The radiance field prediction network predicts RGB colors $\mathbf{c}(\bx)$ and density values $\sigma(\bx)$ from canonicalized points. We encode points $\bx$ with 8 sine components and linearly increase contribution of a single component in $\attributes(\bx)$ and $\bbeta(\bx)$ from 1k to 11k step. Per-point predicted predicted attributes $\attributes(\bx)$ and lifted latent code $\bbeta(\bx)$ are masked by a mask predicted from the masking network depicted in \cref{fig:masknet}. The final linear layer takes additional per-image learnable appearance code $\psi$ to account for any visual variations that cannot be explained by the rest of the framework (\eg changes in lighting). The code can discarded during evaluation. The same layer is additionally conditioned on the positionally encoded view directions. We activate the color output with a sigmoid function.
    }
    \label{fig:nerf}
\end{figure}

We present architecture of: canonicalizer $\Canonicalizer$ in \cref{fig:warping-field}, attribute map $\AttributeNet$ in \cref{fig:attribute-net}, hypermap $\hypermap$ in \cref{fig:hypermap}, per-attribute hypermap in \cref{fig:hypermap-alpha}, mask prediction network in \cref{fig:masknet} and the rendering network in \cref{fig:nerf}. Each network contains only fully connected layers. Hidden layers use ReLU activation function. Colors of figures correspond to colors of blocks in \cref{fig:laced-implicit}.

\section{Additional qualitative results}
See attached video clip for more qualitative results.

\section{Failure Cases}
\begin{figure}
    \centering
    \includegraphics[width=\linewidth]{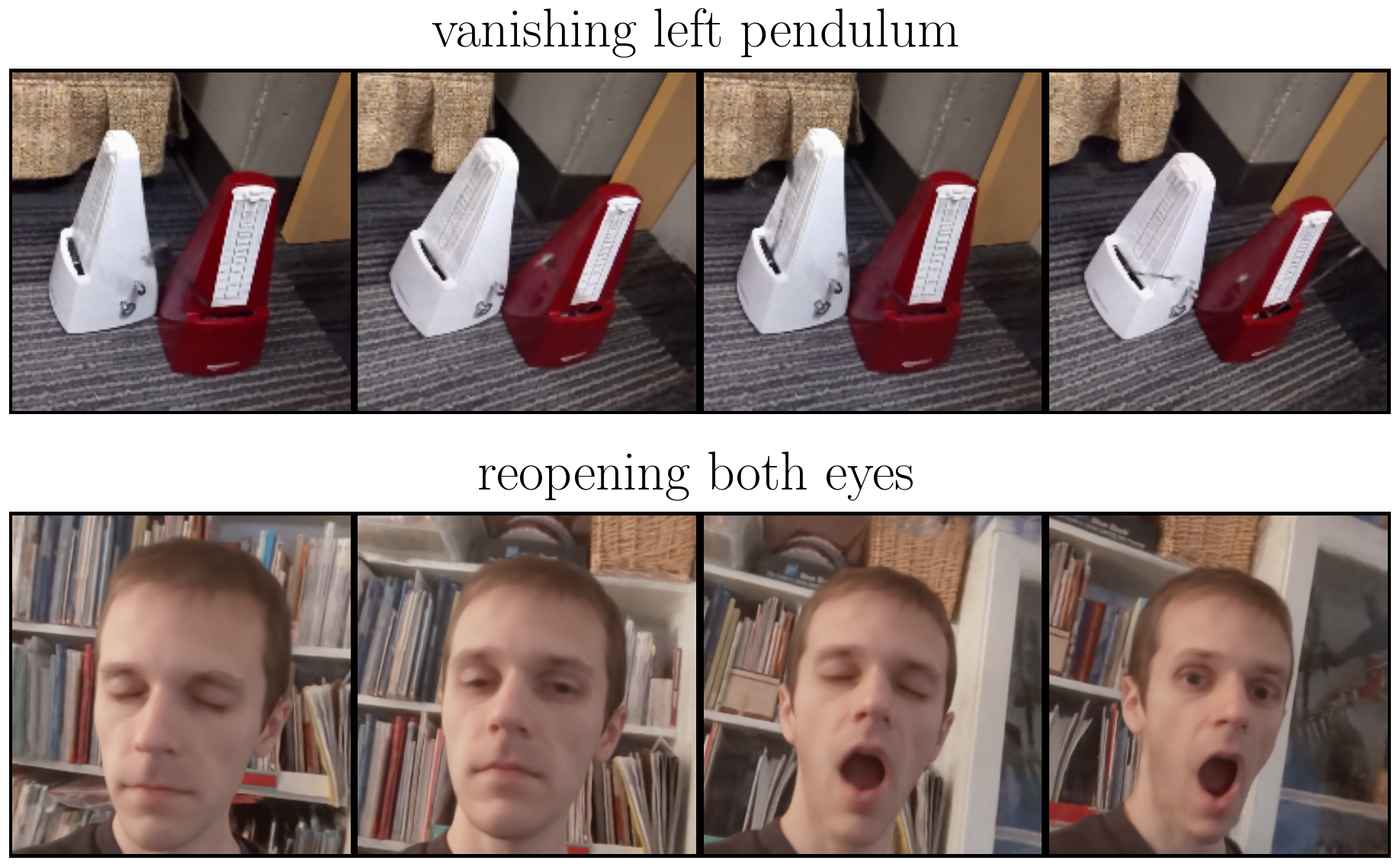}
    \caption{{\bf Failure cases --}
         Our model may learn spurious interpolations for controlled elements that occupy little space in the image and with insufficient/careless annotations.
         For the metronome, due to the fast motion of the pendulum and its specularity, without careful annotation our method may simply learn its motion blur or sometimes even completely ignore the pendulum.
         In the face example, this may result in the eye blinking multiple times while interpolating between the attribute values of $-1$ and $1$.
         Both cases are preventable with more careful annotations and by annotating more frames.
    }
    \label{fig:failure-cases}
\end{figure}
We identify two modes of failure cases in our approach and present them in \cref{fig:failure-cases}. 
In some cases with particular mask annotations, our model can struggle
with controlling elements that occupy small space in the image.
The problem is especially visible for controlling pendulum movement or opening and closing eyes.
In the former, pendulum disappears and reappears in different places.
In the latter, the control of eyes is periodic and there are two distant values in $[-1, 1]$ that produce opening eyes. 
While with careful annotations we noticed that the problem is mostly preventable, this problem may occur in practice.

\clearpage

\end{document}